\documentclass{article}
\usepackage{iclr2026_conference,times}
\iclrfinalcopy
\usepackage{amsmath,amssymb,bm}
\usepackage{booktabs,multirow}
\usepackage{graphicx}
\usepackage[colorlinks=true,linkcolor=blue,citecolor=blue,urlcolor=blue]{hyperref}
\usepackage{xcolor}
\usepackage{caption}
\newcommand{\method}{PTB-Search}
\newcommand{\near}{\textit{near}}
\newcommand{\solve}{\textit{solve}}
\newcommand{\acc}{Acc$_{0.1}$}
\newenvironment{principle}[1]{\medskip\noindent\textbf{#1.}\itshape}{\medskip}

\title{Dictionaries, Not Darwin:\\ Set-Level Selection Beats LLM Evolution\\ in Scientific Equation Discovery}

\author{Pan Li\\Independent Researcher\\\texttt{lipan231@mails.ucas.ac.cn}\\\url{https://github.com/panli0/ptb-search}}
\date{}

\begin{document}
\maketitle

\vspace{-0.75em}
\begin{center}
  \includegraphics[width=0.93\linewidth]{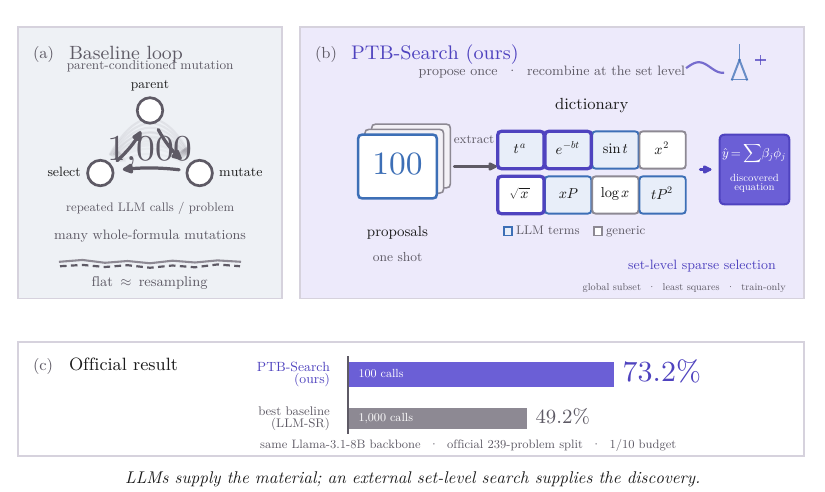}
  \vspace{-0.85em}
  \captionof{figure}{\textbf{Teaser.} A matched-budget audit shows that parent-conditioned LLM ``evolution'' behaves like resampling; \method{} keeps the useful part---proposal material---and replaces generations with one train-only, set-level recombination over a term dictionary, reaching $73.2\%$ official \acc{} at one tenth of the standardized call budget.}
  \label{fig:concept}
\end{center}
\vspace{-0.55em}

\begin{abstract}
Large language models (LLMs) are widely deployed as evolutionary engines for scientific and algorithmic discovery: generate candidates, select winners, feed them back as parents, and repeat. We audit this premise in the setting where such loops should have measurable room to improve: componentized scientific expression discovery. Continuous equation discovery is the central case: finite samples make structure underdetermined, interpolation is easy, and the useful object is a reusable expression rather than a one-off string. Under strictly matched LLM-call budgets, the equation loop is sampling, not evolving: parent-conditioned evolution is indistinguishable from fresh independent sampling (median OOD NMSE $0.045$ vs.\ $0.049$), explicitly instructed multi-parent crossover is worse ($0.200$), final success is predicted by initial proposals (AUC $0.959$), and three distinct iteration schemes fail to add solved problems. Operationally, the loop reduces to what it actually produces: a \emph{dictionary} of candidate terms.

We turn that reduction into a method. \method{} extracts terms from independent proposals and performs one global, train-only, set-level sparse recombination. Its selector follows an identifiability principle: underdetermined training data identifies the joint behavior of term \emph{sets}, not per-term credit. On identical dictionaries with zero additional LLM calls, set-level selectors solve $165$--$169$ of 717 cells where single-term reductions, including marginal-influence credit used by concurrent work, solve only $74$--$78$ (all cross-family comparisons $p<10^{-73}$). On the official 239-problem LLM-SRBench split, \method{} reaches $73.2\%$ numeric accuracy with Llama-3.1-8B and $77.0\%$ with a single-seed DeepSeek-V4 anchor, versus $49.2\%$ for the best reported baseline, while using one tenth of the standardized budget. A program-domain audit supplies a heterogeneous stress test: free-form Python priority functions are not the native input type of \method{}, yet even there the tested LLM lineage loop does not make generation count the active ingredient. The paper's positive claim is therefore intentionally broad but structured: wherever an LLM exposes reusable components and data can score their joint behavior, discovery should be externalized to proposal banks and set-level selection rather than entrusted to prompt lineage.
\end{abstract}

\section{Introduction}
\label{sec:intro}

From program synthesis to equation discovery, a standard architecture has emerged for LLM-driven scientific search: a language model proposes candidate solutions, a scorer ranks them, and top candidates are fed back as parents for the next round of LLM mutation~\citep{funsearch,llmsr,shinka}. The architecture borrows the vocabulary of evolution---populations, parents, generations---and with it an implicit empirical claim: that the loop \emph{compounds}, so that later generations succeed where the same number of independent samples would not. This claim is almost never tested against the null it must beat.

We make that null the object of study. The first domain is continuous equation discovery, chosen deliberately rather than as a toy proxy for science. With finite observations, a continuous formula is structurally underdetermined: many expressions can interpolate the training grid, while only some encode reusable behavior that survives held-out domains. If an LLM evolution loop truly accumulates scientific refinement, this is exactly the kind of setting where it should beat a matched budget of independent proposals.

It does not. Holding the proposer, prompt format, and LLM-call budget fixed on LLM-SRBench-style equation discovery~\citep{llmsrbench}, parent-conditioned proposals match fresh independent sampling (median OOD NMSE $0.045$ vs.\ $0.049$; identical strict-solve counts). Handing the model two strong parents and instructing it to combine their best components makes results \emph{worse} ($0.200$), at $98.7\%$ proposal validity, so parsing failure is not the explanation. Whether a run ultimately succeeds is predicted by the quality of its initial independent proposals with AUC $0.959$, and lineage tracking shows that useful components are only weakly heritable. Three distinct iteration schemes---naive, credit-filtered, and guarded population-level---do not reliably add solved problems across four generations, and the naive scheme is destructive (strict solves $50\to38$). The loop is not accumulating scientific structure; it is resampling with extra steps.

If the loop's dynamics do not supply the discovery, what does? Dissecting the runs shows that the LLM's contribution is \emph{material}: across a problem's independent proposals, the right terms are often present far more often than any single proposal is right. This licenses the reduction that organizes the paper: the entire evolutionary apparatus is, operationally, a \emph{dictionary generator}, and the open design question---the one the loop fails to answer---is selection over the dictionary.

Selection over an LLM-generated dictionary is not a matter of taste; it is governed by identifiability. Training data in this regime is underdetermined: many term subsets fit the training grid almost equally well, so any statistic attached to an \emph{individual} term---marginal influence credit, selection frequency under subsampling, agreement with the ground-truth term set---carries little signal. Indeed, a well-constructed per-term credit correlates with a term's functional contribution (Spearman $0.26$--$0.63$ across families) yet fails to identify ground-truth terms (AUC $0.42$--$0.44$): the data identifies the joint behavior of a candidate \emph{set}, not the attribution of its members. Selectors that score sets jointly should therefore dominate selectors that reduce to per-term statistics. Our largest experiment confirms this as a two-family separation: on 717 cells with \emph{identical} dictionaries and zero additional LLM calls, three set-level joint selectors solve $165$--$169$ problems while two single-term reductions---marginal credit, the mechanism of concurrent work~\citep{igsr}, and stability frequency---solve $78$ and $74$, both landing significantly \emph{below} the frozen baseline they were meant to refine ($p<10^{-37}$).

These two findings assemble into a deliberately simple method. \method{} samples independent LLM proposals once; extracts their additive terms, pooled with generic primitives, into a per-problem dictionary; runs one global, train-only, set-level sparse subset search---best-subset selection with least-squares coefficients, the estimator family sparse dictionary regression has studied for a decade~\citep{sindy}; and stops. No generations. Every component is forced by a collapse ablation (\S\ref{sec:method}), and the architecture is evaluated where it counts: on the official 239-problem LLM-SRBench split, \method{} reaches $73.2\%$ \acc{} with a Llama-3.1-8B backbone and $77.0\%$ with a single-seed DeepSeek-V4 anchor, against $49.2\%$ for the best reported baseline, at one tenth of the standardized call budget (\S\ref{sec:official}). That a weak open model and a strong API model land within four points of each other on the same problems is a stress test of the thesis: once the dictionary contains the right terms, the selector carries a substantial part of discovery, and proposer strength alone is not the dominant driver.

The second domain, online bin-packing program synthesis, is included for a different purpose: it is a deliberately heterogeneous stress test. Free-form Python priority functions are not additive dictionaries, and \method{} is not proposed as a general program-synthesis algorithm. That is exactly why the task is useful: if the evolutionary story were broadly correct, this editable program domain should be friendlier to lineage-conditioned improvement than continuous equations. Instead, easy program instances again tie fresh sampling; on a harder DeepSeek grid, the positive signal attaches to retained external state and a crossover probe rather than to parent-conditioned generation count. We therefore use programs to delimit the method's scope and to rule out an overbroad ``LLM evolution'' interpretation, while the method claim itself remains about componentized discovery spaces where candidate material can be retained and selected jointly (\S\ref{sec:programs}).

Our contributions:
\begin{enumerate}
  \item \textbf{Audit.} A pre-registered, budget-matched, leakage-audited audit showing that the evolution loop in LLM equation discovery is proposal-dominated: fresh $\approx$ vanilla $\approx$ explicit crossover; initial-proposal quality decides success (AUC $0.959$); heritability is weak; three iteration schemes are flat or destructive (\S\ref{sec:audit}). A heterogeneous program-domain audit reaches the same negative conclusion about generation count: the parent-conditioned lineage loop is not the active ingredient, even in a domain where editing is plausible (\S\ref{sec:programs}).
  \item \textbf{Method.} \method{}, a single-generation architecture for componentized scientific expression discovery---independent proposals $\to$ term extraction $\to$ one global set-level sparse recombination---whose every component is justified by a collapse ablation (\S\ref{sec:method}), and which reaches $73.2\%$ \acc{} (Llama-3.1-8B) and $77.0\%$ \acc{} (DeepSeek-V4) on the official benchmark at one tenth of the standardized budget. The cross-backbone stability itself corroborates that once enough reusable components are exposed, external selection is a load-bearing part of discovery (\S\ref{sec:official}).
  \item \textbf{Selector principle.} Evidence that selection signals must live at the set level: a two-family separation ($165$--$169$ vs.\ $74$--$78$ solves of 717; every cross-family pair at $p<10^{-73}$), robust across proposers and dictionary capacities, with an identifiability-based account of why per-term reduction fails (\S\ref{sec:selectors}).
  \item \textbf{Mechanism accounting.} A failure-mode decomposition of the remaining error (selection vs.\ availability vs.\ composition), and a complexity-guard mechanism experiment showing that apparent one-layer nonlinear-composition gains are bought with $8\times$ formula length---hence excluded by the guard, in agreement with the decomposition's near-zero composition gap (\S\ref{sec:remaining}).
\end{enumerate}

\section{Setup and protocol}
\label{sec:setup}

\paragraph{Task and metrics.} We use LLM-SRBench equation discovery~\citep{llmsrbench}: given training samples of an unknown ground-truth formula, produce a symbolic expression. The official benchmark contains 239 problems---111 LSR-Transform (transformed Feynman physics) and 128 LSR-Synth across chemistry (36), biology (24), physics (43), and material science (25). Official metrics are symbolic accuracy (SA), numeric accuracy \acc{}, and NMSE; the official Table-1 aggregation is the \emph{median} NMSE across problems, the official protocol standardizes discovery methods to 1{,}000 LLM calls per problem, and the benchmark supplies OOD samples for the 128 LSR-Synth problems. When reporting on the official split we follow Table-1 style exactly: per-problem mean over our three seeds, then median across problems. For method-development audits we additionally report median OOD NMSE and two thresholded counts, \near{} (OOD NMSE $<10^{-3}$) and strict \solve{} ($<10^{-6}$), computed strictly post-hoc.

\paragraph{Grids.} Method development uses two DeepSeek splits, both frozen before the official evaluation: a development split (25 problems $\times$ seeds 0--2 $=$ 75 cells) used to design and fix every hyperparameter, and a blind split (30 problems $\times$ seeds 0--2 $=$ 90 cells) evaluated only after the method was frozen. The official 239-problem split is then run end to end with \emph{two} proposers under the frozen method: a local Llama-3.1-8B benchmark (239 problems $\times$ seeds 0--2 $=$ 717 cells) and a DeepSeek-V4 anchor (239 problems, seed 0, $23{,}900$ fresh proposals). A cell is a problem--seed pair. The DeepSeek official row is a single-seed anchor, reported as such; the Llama benchmark is the three-seed grid.

\paragraph{Budget fairness and leakage contract.} Every LLM arm uses the same per-cell completion budget (100 completions); external search arms additionally use CPU search, declared as part of the method. Selection, dictionary construction, hyperparameters, and early stopping use training data only (internal 70/30 splits and a train-side guard fold); OOD data and ground-truth formulas are never touched by any method component. Comparisons are paired at the cell level within the same proposer; cross-proposer pairing is disallowed. Every experiment carries pre-registered interpretation branches written before execution, and we report the branch that occurred. Leakage audits, config hashes, and completeness checks accompany every result directory (App.~\ref{app:audit-artifacts}).

\section{Auditing the evolution loop}
\label{sec:audit}

\begin{table}[t]
\centering\small
\caption{Budget-matched audit, DeepSeek development split (75 cells, 100 LLM calls per cell in every arm). Evolution does not beat independent sampling; instructed crossover is worse; the externalized architecture is seven orders of magnitude better in median OOD NMSE. Validity is the fraction of parseable proposals.}
\label{tab:audit}
\begin{tabular}{lcccc}
\toprule
Arm & Median OOD NMSE & \near{} & \solve{} & Validity \\
\midrule
Fresh independent sampling      & $4.91\times10^{-2}$ & 23 & 16 & 0.995 \\
Parent-conditioned evolution    & $4.47\times10^{-2}$ & 21 & 16 & 0.940 \\
Instructed multi-parent crossover & $2.00\times10^{-1}$ & 18 & 11 & 0.987 \\
\method{} (frozen)              & $\bm{4.7\times10^{-9}}$ & \textbf{57} & \textbf{50} & 1.0 \\
\bottomrule
\end{tabular}
\end{table}

\begin{figure}[!t]
  \centering
  \includegraphics[width=0.92\linewidth]{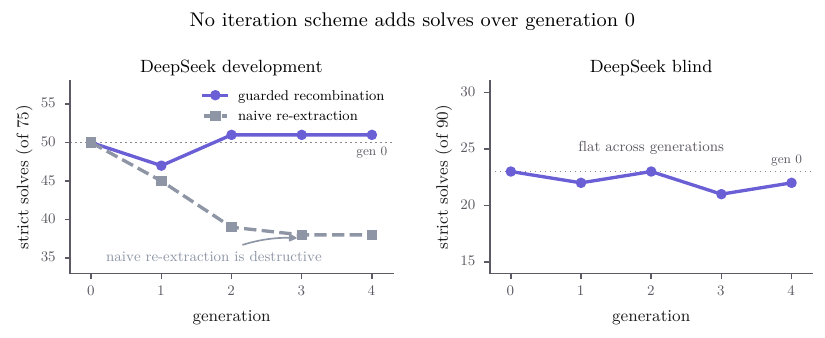}
  \caption{\textbf{Generations do not add solves.} Guarded population-level recombination is flat across generations, while naive re-extraction destroys solves; the loop has no residual cumulative work after the dictionary has been built.}
  \label{fig:generations}
\end{figure}

\paragraph{D1: The loop does not beat its null.} Under identical budgets and proposer, parent-conditioned evolution matches fresh sampling ($0.045$ vs.\ $0.049$ median OOD NMSE; strict solves 16 vs.\ 16; Table~\ref{tab:audit}). Instructing the model to recombine---multiple top parents in the prompt, with explicit directions to combine their best components---is \emph{worse} ($0.200$) despite near-perfect proposal validity ($0.987$), which rules out parsing failure as the explanation. The LLM, even when asked directly, does not recombine.

\paragraph{D2: Initial proposals decide.} Whether a run reaches a good final equation is predicted by the quality of its independently sampled initial proposals with AUC $0.959$; conditioning later calls on selected parents adds no measurable lift. (Initial-proposal quality is the best post-hoc OOD NMSE among a cell's round-0 proposals, used only for audit analysis, never for selection.)

\paragraph{D3: Weak heritability; per-term credit is unidentifiable.} Lineage tracking shows useful terms survive across generations only weakly, with accumulation of low-credit terms as the dominant dynamic. A three-factor per-term credit (selection frequency $\times$ standardized coefficient $\times$ drop-one loss increase) correlates with post-hoc functional contribution---Spearman $0.26$--$0.63$ against drop-one OOD change across families---yet fails to identify ground-truth terms: AUC $0.44$ (95\% CI $[0.36,0.53]$) on the development split and $0.42$ ($[0.35,0.51]$) on the blind split, with no family exceeding chance reliably. The credit is measuring \emph{functional} contribution inside one fitted set, not \emph{syntactic} truth---the first sign that per-term attribution is not what the data determines.

\paragraph{D4: Generations never pay.} Three independent iteration schemes on top of externalized recombination---naive re-extraction, credit-filtered heritable recombination with early stopping, and guarded population-level recombination---are flat in strict solves across four generations on both DeepSeek splits (guarded scheme: $50,47,51,51,51$ on the development split; $23,22,23,21,22$ blind). The naive scheme decays ($50\to38$), and the credit-filtered scheme inflates median formula length ($270\to523$) with no OOD gain (Fig.~\ref{fig:generations}). Unguarded, iteration is not merely useless; it is destructive.

\begin{principle}{Principle 1 (materials, not trajectories)}
The value of LLM proposals for equation discovery is which terms exist across them, not how formulas are rewritten between them. An architecture should harvest proposals as material and place no weight on proposal-to-proposal dynamics.
\end{principle}

\section{\method{}: dictionary regression on LLM proposals}
\label{sec:method}

Viewed statistically, the architecture that Principle~1 licenses is classical. Let the training data be $\{(\bm{x}_i, y_i)\}$. Independent LLM proposals for the problem are parsed and their additive terms canonicalized into a dictionary $D=\{\phi_1,\dots,\phi_p\}$ of candidate basis functions, pooled with a small set of generic primitives. The hypothesis class is sparse linear combinations $\hat y = \beta_0 + \sum_{j\in S}\beta_j \phi_j(\bm{x})$ with $|S|\le 6$; given a support $S$, coefficients are least squares, so the estimation problem given the dictionary is best-subset selection---the estimator family of sparse dictionary regression~\citep{sindy}. The LLM's entire role is to make $D$ contain the right functions; every other evolutionary operator is externalized:

\begin{enumerate}
  \item \textbf{Variation (LLM, once).} $n{=}100$ independent proposals with a fixed prompt; no parent conditioning.
  \item \textbf{Inheritance (extraction).} Terms outlive their parent formulas by entering $D$; this is where heritability actually lives.
  \item \textbf{Recombination $+$ selection (external, set-level).} One global sparse subset search over $D$: beam search over supports scored by internal-split train NMSE with least-squares fitting. The full selector adds pairwise-synergy-aware scoring, stability-generated candidate supports, and a train-side guard fold used only to break statistical ties; all hyperparameters are chosen on internal train splits.
  \item \textbf{Stop.} No generations (D4).
\end{enumerate}

\paragraph{Design forced by ablations.} The method has no component that survives on faith. Restricting extraction to a single proposal collapses the development split from $4.7\times10^{-9}$ to $2.2\times10^{-2}$ median OOD (fresh level) and the blind split to $0.103$ (5 strict solves vs.\ 23): cross-proposal pooling is the point. Dense fitting of the full dictionary without subset selection is uniformly worse (blind $2.1\times10^{-3}$ vs.\ $9.5\times10^{-4}$; 19 vs.\ 23 solves): selection is necessary, not decorative. A local-edit-only operator on a single proposal lands strictly between single-proposal and full \method{} on both splits, locating the gain in cross-proposal recombination rather than local refinement. Transplanting terms across problem families fails outright (0 \near{}, 0 \solve{}, validity $0.133$): the dictionary's value is problem-compatible material, not arbitrary extra basis functions. Replacing set-level search by per-term ranking forfeits half the solves (\S\ref{sec:selectors}); adding generations gives no reliable gains in these runs (D4). The support-size cap ($|S|\le 6$) is not a free prior but one arm of the same complexity guard that governs nonlinear layers: the mechanism experiment of \S\ref{sec:remaining} shows that relaxing the budget buys formula length, not held-out accuracy, so the cap is a scope rule the data endorses rather than an assumption the method rests on.

\paragraph{What the LLM contributes.} Term-source ablations position the LLM precisely: on the development split, LLM-only / generic-only / pooled dictionaries give median OOD $3.5\times10^{-6}$ / $2.7\times10^{-5}$ / $4.7\times10^{-9}$ and strict solves $36$ / $35$ / $50$; on the blind split, $4.9\times10^{-3}$ / $1.9\times10^{-3}$ / $9.5\times10^{-4}$ and $15$ / $24$ / $23$. LLM terms are complementary, problem-compatible priors on top of a generic-primitive dictionary that is itself a strong baseline---a fact we state rather than obscure, because it calibrates exactly how much of the system's intelligence resides in proposal versus selection.

\section{Official-benchmark results}
\label{sec:official}

\begin{table}[t]
\centering\small
\caption{Official LLM-SRBench comparison. Baseline rows are problem-count-weighted composites of the official Table-1 Llama-3.1-8B rows (each composite reproduces exactly from the per-domain entries); the official protocol standardizes discovery methods to $1{,}000$ LLM calls per problem, while \method{} uses $100$. \acc{} is a problem fraction and is therefore directly comparable across rows; the median-NMSE column is left blank for baselines because the official table reports only per-domain NMSE, with no comparable all-239 median, and we do not manufacture one. \dag: SA for \method{} is evaluated under the official-judge pipeline on a train-only simplified readout and is reported as a caveated diagnostic, not a headline (\S\ref{sec:official}, ``Symbolic readout''); the DeepSeek row is a seed-0 anchor.}
\label{tab:official}
\begin{tabular}{llcccc}
\toprule
Method & Backbone & SA (\%) & \acc{} (\%) & Median NMSE (ours) & Calls/prob. \\
\midrule
Direct prompting & Llama-3.1-8B & 1.7 & 0.8 & --- & 1{,}000 \\
SGA~\citep{sga}  & Llama-3.1-8B & 1.3 & 3.4 & --- & 1{,}000 \\
LaSR~\citep{lasr} & Llama-3.1-8B & 4.6 & 38.4 & --- & 1{,}000 \\
LLM-SR~\citep{llmsr} & Llama-3.1-8B & 19.7 & 49.2 & --- & 1{,}000 \\
\midrule
\method{} (ours) & Llama-3.1-8B & --\,\textsuperscript{\dag} & \textbf{73.2} & $1.30\times10^{-3}$ & \textbf{100} \\
\method{} (ours) & DeepSeek-V4 & --\,\textsuperscript{\dag} & \textbf{77.0} & $9.24\times10^{-5}$ & \textbf{100} \\
\bottomrule
\end{tabular}
\end{table}

\paragraph{Protocol.} We evaluate on the official 239-problem split with the official evaluation code and Table-1 aggregation (per-problem mean over available seeds, then median across problems). Baseline numbers are the official benchmark's Llama-3.1-8B rows, recombined into all-239 composites weighted by problem count; each composite reproduces the reported per-domain entries exactly (App.~\ref{app:official-composites}). We run the identical frozen method with two proposers---the local Llama-3.1-8B grid (three seeds) and a DeepSeek-V4 anchor (seed 0, $23{,}900$ fresh proposals)---and no component of \method{} touches OOD data, ground-truth formulas, or the judge in either run.

\paragraph{Results.} \method{} reaches $73.2\%$ \acc{} with the Llama backbone against $49.2\%$ for the best reported baseline, at one tenth of the standardized call budget (Table~\ref{tab:official}). Run unchanged on the DeepSeek-V4 proposer it reaches $77.0\%$ in a single-seed anchor: the two proposer runs---a small open model and a strong API model, differing by orders of magnitude in capability---land within four points of each other on the same 239 problems. This small cross-proposer gap, read only as a corroborating anchor rather than a multi-seed benchmark claim, is a useful corroborating observation for the paper's thesis. If the loop's mutation dynamics or the proposer's raw strength were the only drivers of discovery, a four-point spread across such different models would be less expected; under the dictionary-and-selector account, the result is consistent with both proposers surfacing enough problem-compatible terms for the same selector to recombine. On the 128 LSR-Synth problems, where the benchmark supplies OOD data, the Llama run reaches $95.3\%$ ID and $71.9\%$ OOD \acc{} with median OOD NMSE $1.59\times10^{-5}$; material science is solved essentially outright ($100\%$ ID and $88.0\%$ OOD \acc{}, median OOD NMSE $1.14\times10^{-7}$). The transformed-Feynman family (LSR-Transform) is where the method is weakest ($47.8\%$ ID \acc{} on Llama, $55.0\%$ on DeepSeek): compact transformed closed forms stress dictionary coverage, consistent with the family-level analysis of \S\ref{sec:selectors}, and we name it as the benchmark's designated hard regime rather than smoothing over it. We report the all-239 median NMSE for our own runs, clearly labeled, and make no NMSE comparison against baselines the official table does not support.

\paragraph{Symbolic readout.} \method{}'s native readout is a sparse additive model with fitted coefficients. Symbolic accuracy judges equivalence to a compact closed form, so an \emph{unsimplified} sparse readout is judged inequivalent nearly by construction, independently of numeric fidelity---material science makes the point sharply ($100\%$ ID \acc{} with near-zero raw SA). We therefore report SA under a train-only symbolic post-processing of the readout---guard-consistent term pruning, coefficient snapping, canonical simplification---which measures the symbolic content of the discovered model rather than the verbosity of its raw parameterization; replacement-judge SA is reported only as a caveated diagnostic while the official GPT-4o judge is unavailable.

\section{Selector analysis: selection lives at the set level}
\label{sec:selectors}

\begin{table}[t]
\centering\small
\caption{Selector comparison. Identical dictionaries and zero additional LLM calls within each group; selectors differ only in where their signal lives. Both single-term reductions are significantly worse than the frozen set-level baseline; set-level joint selectors match or beat it.}
\label{tab:selectors}
\begin{tabular}{llccc}
\toprule
Group & Selector (family) & Median OOD & \near{} & \solve{} \\
\midrule
\multirow{4}{*}{\shortstack[l]{DeepSeek dev\\(75 cells)}}
 & MarginalTop (single-term)   & $2.98\times10^{-8}$ & 47 & 40 \\
 & StabilityTop (single-term)  & $2.00\times10^{-8}$ & 48 & 42 \\
 & \method{} frozen (set-level beam) & $4.7\times10^{-9}$ & 57 & 50 \\
 & \method{} Full (set-level)  & $\bm{4.71\times10^{-9}}$ & \textbf{58} & \textbf{50} \\
\midrule
\multirow{4}{*}{\shortstack[l]{DeepSeek blind\\(90 cells)}}
 & MarginalTop (single-term)   & $9.12\times10^{-3}$ & 33 & 10 \\
 & StabilityTop (single-term)  & $7.36\times10^{-3}$ & 36 & 12 \\
 & \method{} frozen (set-level beam) & $9.50\times10^{-4}$ & 45 & 23 \\
 & \method{} Full (set-level)  & $9.50\times10^{-4}$ & 45 & \textbf{25} \\
\midrule
\multirow{6}{*}{\shortstack[l]{Llama-3.1-8B\\full benchmark\\(717 cells)}}
 & Fresh sampling (no selector) & $4.23\times10^{-1}$ & 28 & 5 \\
 & MarginalTop (single-term)   & $1.23\times10^{-1}$ & 174 & 78 \\
 & StabilityTop (single-term)  & $1.29\times10^{-1}$ & 173 & 74 \\
 & \method{} frozen (set-level beam) & $2.75\times10^{-2}$ & 239 & 135 \\
 & CompatBeam (set-level)      & $2.08\times10^{-2}$ & 261 & 167 \\
 & \method{} Full (set-level)  & $\bm{2.08\times10^{-2}}$ & \textbf{262} & \textbf{169} \\
\bottomrule
\end{tabular}
\end{table}

\begin{figure}[t]
  \centering
  \includegraphics[width=0.82\linewidth]{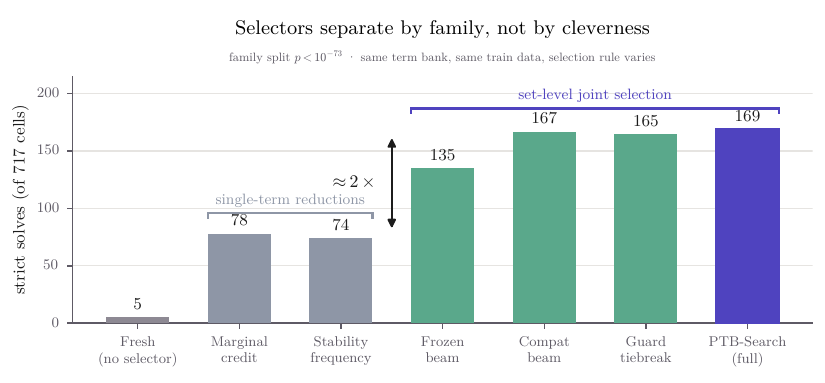}
  \caption{\textbf{Selector family, not clever scoring, decides the result.} On the same 717-cell Llama grid and identical term dictionaries, every set-level selector beats every single-term reduction by roughly $2\times$ in strict solves; both reductions fall below the frozen set-level baseline.}
  \label{fig:families}
\end{figure}

\paragraph{The two-family separation.} On 717 cells, the three set-level joint selectors solve $165$--$169$ problems; the two single-term reductions solve $74$--$78$ (Table~\ref{tab:selectors}, Fig.~\ref{fig:families}). Every cross-family paired comparison is significant at $p<10^{-73}$; the full selector beats marginal credit by $+91$ solves (95\% CI $[75,109]$, $p=1.4\times10^{-85}$). Critically, \emph{both} reductions---marginal influence and stability frequency, two very different statistics---fail identically, and both are significantly worse than the frozen set-level baseline they were meant to refine ($-57$ and $-61$ solves; $p=2.3\times10^{-59}$ and $4.2\times10^{-38}$). The failure axis is not the choice of per-term statistic; it is per-term reduction itself, consistent with D3's identifiability analysis. All sign-tests exclude exact ties and are computed at the paired cell level; the cross-family separations sit at $p<10^{-73}$, so they survive any multiple-comparison correction over the handful of selector contrasts we run by many orders of magnitude.

\paragraph{Refinements pay where the proposer is weak.} The full selector beats the frozen baseline on 717 cells by $+23$ \near{} (CI $[11,35]$) and $+34$ \solve{} (CI $[23,46]$, $p=2.7\times10^{-7}$), with gains concentrated in material science ($36\to55$ solves), chemistry ($38\to45$), and physics ($29\to34$), and the hard LSR-Transform family flat ($2\to2$). On DeepSeek, where the frozen baseline already sits near its ceiling, the blind gain is a modest $+2$ solves. The pattern is the useful one: set-level refinements matter most when the proposer is weak and the dictionary is noisy---precisely the regime where selection has to carry the system.

\begin{figure}[t]
  \centering
  \includegraphics[width=0.84\linewidth]{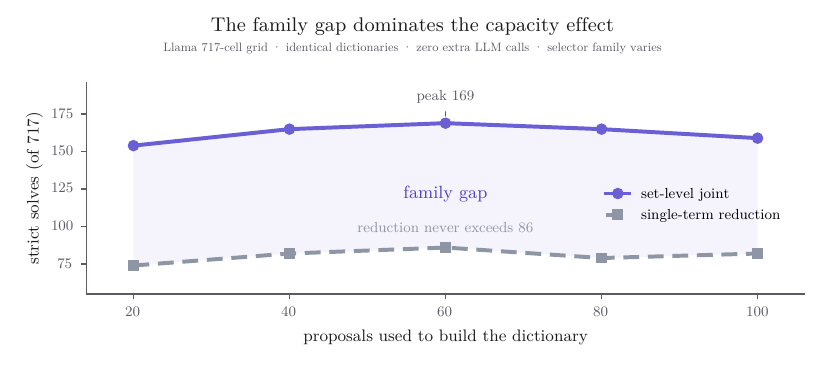}
  \caption{\textbf{More material does not rescue the wrong selector.} Across 20--100 proposal roots, the set-level family dominates the single-term family at every capacity; the selector family gap is larger than the material-quantity effect.}
  \label{fig:capacity}
\end{figure}

\paragraph{Capacity analysis.} Sweeping the number of proposals used to build the dictionary from 20 to 100 (Fig.~\ref{fig:capacity}): the set-level family peaks mildly at 60 proposals on the Llama grid (169 solves) and degrades only slightly at 100 (159); the reduction family never exceeds 86 solves at any capacity. On DeepSeek blind, the reduction selector \emph{loses} solves as material grows ($11\to5$ from 20 to 100) while the set-level selector is stable ($25\to23$). Set-level selection is also the better capacity control, tolerating larger, noisier dictionaries.

\begin{principle}{Principle 2 (set-level identifiability)}
Underdetermined training data identifies the joint behavior of a term set, not the attribution of individual terms. Selection signals must therefore be computed on candidate sets; any reduction to per-term statistics discards the part of the signal the data actually determines.
\end{principle}

\begin{figure}[t]
  \centering
  \includegraphics[width=0.98\linewidth]{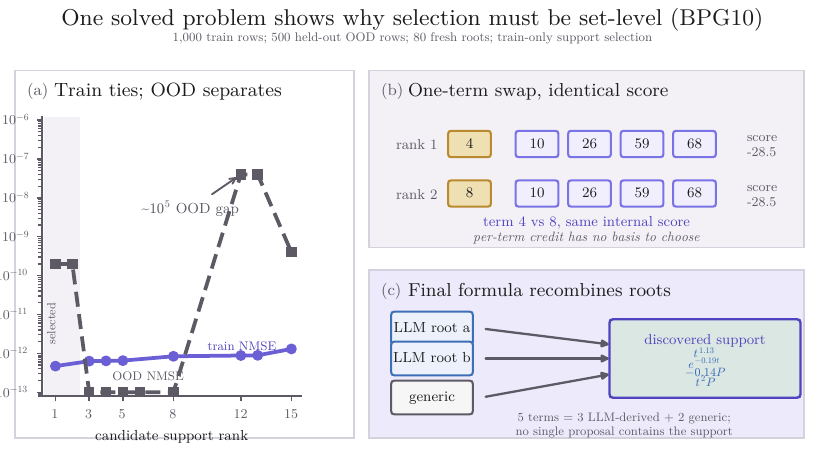}
  \caption{\textbf{One solved problem makes set-level identifiability visible.} (A) Many candidate supports are essentially tied on training NMSE, but their held-out OOD errors spread across roughly five orders of magnitude. (B) The two best supports differ by one swapped term at identical internal score, so per-term credit cannot choose between them. (C) The final support recombines terms from multiple LLM roots plus generic primitives; no single proposal contained the selected set.}
  \label{fig:interpretability}
\end{figure}

\paragraph{The principle, seen in one problem.} Figure~\ref{fig:interpretability} makes the abstraction physical on a single solved case. The visible flat valley---many supports tied on train, spread across five OOD decades---is precisely the underdetermination that Principle~2 names, and the one-term swap between the two best supports at identical score is precisely why a per-term credit cannot recover the answer: the information lives in which \emph{set} extrapolates, and the training grid does not carry it. That the winning formula is assembled from terms sourced across multiple independent proposals, none of which held the full set, is the constructive side of the same fact---recombination is doing the discovery.

\section{Where the remaining error lives}
\label{sec:remaining}

\begin{figure}[t]
  \centering
  \includegraphics[width=0.5\linewidth]{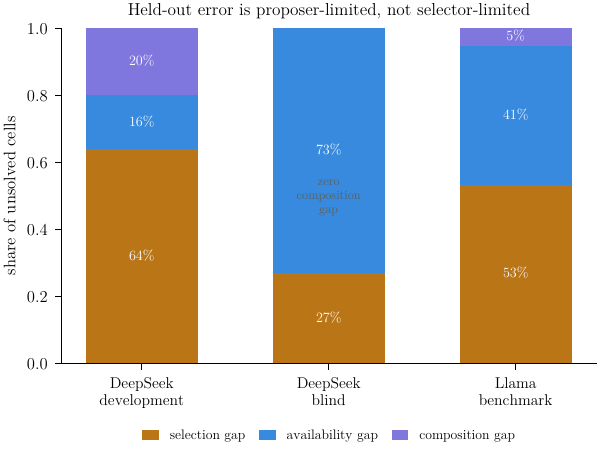}
  \caption{\textbf{The remaining failures are mostly proposer-limited or selector-limited, not composition-limited.} A strictly post-hoc oracle autopsy separates selection, dictionary availability, and composition gaps; no oracle signal is fed back into any method component.}
  \label{fig:autopsy}
\end{figure}

\paragraph{Autopsy.} For every unsolved cell we determine, post-hoc, whether an oracle could have solved it from the existing dictionary (selection gap), whether required terms are absent (availability gap), or whether the true formula's top-level structure exceeds linear composition (composition gap). The decomposition (Fig.~\ref{fig:autopsy}) quantifies each component's theoretical headroom and differs sharply by regime: the development split is selection-dominated ($64\%$), the blind split availability-dominated ($73\%$, with \emph{zero} composition gap), and the Llama benchmark mixed ($53\%$ / $41\%$ / $5\%$). Held-out error is dominated by what the proposer never wrote down---a proposer-capability boundary, not a selector defect.

\paragraph{The complexity guard bounds the method's scope.} The guard also settles, mechanistically, why \method{} is linear in its terms. Extending candidates with one nonlinear wrap or product-of-sums layer raises blind \near{} ($45\to47$) at the price of $8\times$ median formula length ($267\to2075$; parameters $7\to45$), so the pre-registered guard excludes it on nearly all cells---exactly what the autopsy's near-zero composition gap predicts. Apparent composition gains on this benchmark are purchased with complexity, and the guard converts that observation into a scope rule: a component enters the method only if its gains survive the length budget.

\section{A heterogeneous program audit checks the scope of the method}
\label{sec:programs}

The equation-domain method rests on flat componentization: useful fragments can be extracted, retained, and recombined by a train-only set-level selector. That structure also appears beyond polynomial expressions---in symbolic regression, dynamical-system term discovery, physics-inspired libraries, and formula-like heuristic scores---but it does not automatically cover arbitrary programs. We therefore use online one-dimensional bin packing with LLM-proposed Python priority functions (the FunSearch task family) as a stress test rather than as a second application of \method{}. The question is deliberately narrower: does the common parent-conditioned LLM lineage loop become a reliable cumulative mechanism once the search object is an editable program rather than a flat expression?

\paragraph{Setup.} A DeepSeek proposer produces priority heuristics; selection and parent choice use training instances only, and blind instance sets from distinct distributions are evaluated strictly post-hoc. The easy grid uses three near-uniform distributions $\times$ five seeds at a matched budget of $N{=}200$ calls per cell. The hard grid uses three heavier-tailed distributions (OR-style, Weibull, tight) $\times$ three clean seeds at $N{=}400$ (seeds 3--4 were excluded after billing/API errors rather than repaired for this quick readout), comparing fresh independent sampling, parent-conditioned evolution over 20 generations, engine-side external-population evolution over 20 generations, and an instructed two-parent crossover probe at a quarter of the budget. The metric is blind excess-bin rate (lower is better); proposal validity is essentially perfect in every arm ($\ge 0.99$).

\paragraph{Easy instances reproduce the equation-domain null.} At the easy difficulty, no evolutionary arm is significantly better than matched-budget fresh sampling: paired sign-tests give $p{=}0.18$ (parent-conditioned), $p{=}0.75$ (external population), and $p{=}0.34$ (crossover probe), with exact ties in a third of the cells and mean blind-excess differences below $0.2$ on a scale where arm means are ${\approx}6$. Here the loop again shows no significant advantage over sampling.

\begin{figure}[t]
  \centering
  \includegraphics[width=\linewidth]{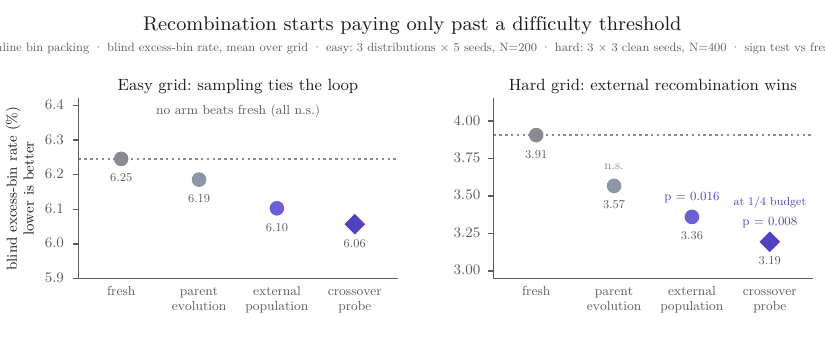}
  \caption{\textbf{Program search reveals the boundary of the account.} Easy bin-packing instances reproduce the sampling null; harder instances reward recombination over an external population, while the parent-conditioned loop remains non-significant. The gain attaches to retained external state, not generation count.}
  \label{fig:program-boundary}
\end{figure}

\paragraph{Hard instances give a scoped signal (Table~\ref{tab:programs}).} When the instance distribution is made harder, the clean 3-seed DeepSeek picture separates: recombination over an external population beats matched-budget fresh sampling on $7$ of $9$ non-tie cells (paired sign-test $p{=}0.016$), and the instructed crossover probe beats it on $8$ of $9$ ($p{=}0.008$) while attaining the best mean blind excess using only a quarter of the budget. Parent-conditioned evolution---the arm that adds generations without an external set to recombine over---remains indistinguishable from fresh ($5/3/1$, $p{=}0.73$). We read the result conservatively. It is evidence that retained external state can matter in this heterogeneous stress test, but it is not a claim that \method{} directly solves free-form program synthesis or that one program-mechanism ranking transfers across proposers.

\begin{table}[t]
\centering\small
\caption{Program domain, hard DeepSeek instances (three heavy-tailed distributions $\times$ three clean seeds, matched budget). Blind excess-bin rate, lower is better; the paired columns test each arm against fresh sampling on the same distribution/seed (ties excluded from the sign-test). This is a scoped boundary readout: external-state recombination helps in this grid, while adding generations to a parent loop does not.}
\label{tab:programs}
\begin{tabular}{lcccc}
\toprule
Arm & Budget & Mean blind excess (\%) & Wins/Losses vs.\ fresh & Sign-test $p$ \\
\midrule
Fresh independent sampling      & $N$   & $3.91$ & --- & --- \\
Parent-conditioned, 20 gen.     & $N$   & $3.57$ & $5/3$ & $0.73$ \\
External-population, 20 gen.     & $N$   & $3.36$ & $7/0$ & $\mathbf{0.016}$ \\
Instructed crossover probe      & $N/4$ & $\mathbf{3.195}$ & $8/0$ & $\mathbf{0.008}$ \\
\bottomrule
\end{tabular}
\end{table}

\paragraph{Reading the two grids together.} The overclaim the paper targets is specifically the belief that \emph{iterating the LLM} accumulates improvement. That belief is not supported in the tested settings: equation generations are flat (D4), easy-program generations tie fresh, and even on hard programs the multi-generation parent loop ties fresh. The program evidence therefore protects the method claim rather than diluting it. \method{} is a method for componentized proposal spaces; the program stress test shows that leaving that regime does not resurrect the simple Darwinian interpretation of the LLM loop.

\begin{principle}{Principle 3 (when one generation suffices, and what a later one would buy)}
When candidate solutions expose reusable components that can be scored jointly, recombination, credit, and selection should be externalized rather than delegated to prompt lineage. Linear additive dictionaries are the cleanest case, but the scope is broader: the requirement is decomposable material plus a data-driven set selector. Free-form programs violate this flat-component assumption, and in the scoped program audit the parent-generation count still does not become the active ingredient.
\end{principle}

\section{Related work}
\label{sec:related}

\paragraph{LLM systems for scientific and algorithmic discovery.}
A growing line of work uses LLMs not merely to answer scientific questions, but to participate in search: propose a formula, program, heuristic, hypothesis, or design; evaluate it with data or a simulator; keep the best candidates; and use them as context for the next round. FunSearch~\citep{funsearch} popularized this pattern for mathematical program discovery; LLM-SR~\citep{llmsr}, LaSR~\citep{lasr}, ShinkaEvolve~\citep{shinka}, and LLM-SRBench~\citep{llmsrbench} instantiate closely related loops for scientific equation discovery. This paper targets the mechanism shared by these systems, not a single benchmark score: when an LLM discovery system appears to improve over generations, is the improvement caused by cumulative LLM evolution, or by a simpler combination of proposal material, archive state, and external selection?

\paragraph{The missing matched-sampling null.}
Most LLM-discovery papers report end-task success but do not isolate the null the loop must beat: spending the same number of LLM calls on independent proposals. That omission matters because LLM proposals are already strong one-shot priors, and scientific discovery tasks often have cheap verifiers but underdetermined search spaces. Recent independent evidence points in the same direction---a genetic algorithm on the same ansatz representation can outperform an LLM loop using far fewer evaluations~\citep{bicycle}; constraining the LLM to a tool role improves architecture-search behavior~\citep{llmtool}; and diversity-oriented hypothesis-search work explicitly worries about collapse under iterative LLM operators~\citep{diversehyp}. We make the matched-sampling null the center of the paper and test it in two domains: continuous scientific equation discovery, where interpolation is easy but structural OOD behavior is hard, and online bin-packing program discovery, where programs can in principle be edited across generations.

\paragraph{Continuous equation discovery and identifiability.}
Continuous equation discovery is a classic scientific-discovery setting, not a convenience benchmark. With finite samples from an unknown law, many expressions can fit the observed grid, while only some capture the reusable structure that extrapolates. This is exactly the setting where a claimed scientific-search mechanism must be audited: the difficulty is not merely finding a low training error, but distinguishing structural terms from interpolation artifacts. Sparse regression and symbolic regression have studied this issue for decades, from free-form law discovery~\citep{schmidt2009distilling} through SINDy~\citep{sindy}, AI Feynman~\citep{udrescu2020aifeynman}, PySR~\citep{pysr}, and unified deep-SR frameworks~\citep{udsr}. \method{} deliberately reuses this older lesson: once LLM proposals expose a correlated dictionary of candidate scientific terms, the load-bearing operation is not another natural-language mutation, but a train-only sparse estimator that scores term \emph{sets} jointly.

\paragraph{Concurrent work: per-term credit.}
IGSR~\citep{igsr} also decouples LLM generation from external selection over proposed basis functions, scoring each term by its marginal influence inside a propose-and-prune search. We view that convergence as evidence that LLM scientific discovery is moving away from pure prompt-lineage evolution. The difference is the identifiability boundary. IGSR's credit is a per-term marginal statistic; our experiments show that the entire family of single-term reductions, implemented on identical dictionaries, sits on the losing side of the boundary, forfeiting roughly half of all solved problems relative to set-level joint search (\S\ref{sec:selectors}). The distinction matters scientifically: under underdetermined data, the observable object is often a behaviorally equivalent \emph{support}, not a uniquely attributable term.

\paragraph{Audits of claimed mechanisms.}
Method papers often owe part of their impact to a mechanism story that later turns out to be too strong. Re-auditing that story under controlled protocols has changed the understanding of several mature areas: pruning gains often came from sparse architectures rather than inherited weights~\citep{liu2019pruning}; MAML's fast adaptation often reduced to feature reuse~\citep{raghu2020rapid}; carefully tuned baselines re-ordered claims in language modeling~\citep{melis2018state}; and evaluation-discipline papers in deep RL and GANs showed how unstable comparisons can overstate novelty~\citep{henderson2018deep,lucic2018gans}. We place LLM scientific discovery in the same tradition. The contribution is not only a better equation-discovery system, but a controlled accounting of what current LLM discovery loops are actually buying.

\section{Discussion}
\label{sec:discussion}

\paragraph{What this says about LLM scientific discovery.}
The paper's main claim is not that LLMs are poor scientific generators. The official-split results say the opposite: a small open model and a strong API model both provide enough problem-compatible material for a fixed selector to beat the reported baseline at one tenth of the standardized call budget. The claim is about where the discovery signal resides. In these systems, the LLM supplies candidate material; the verifier supplies feedback; the archive stores state; and the external selector decides which pieces jointly explain the data. Calling the whole loop ``evolution'' hides this division of labor and encourages the wrong ablation. The important question is not whether later prompts are conditioned on earlier winners, but whether retained material is recombined by a decision rule that the data can identify.

\paragraph{Why continuous science exposes the problem.}
Continuous equation discovery makes the failure mode visible because the task is underdetermined in exactly the scientific sense: many formulas can interpolate finite samples, but only some encode reusable structure. A parent-conditioned LLM loop can easily keep rewriting train-good formulas without accumulating the structural pieces needed for OOD behavior. A dictionary view explains both sides of the result. Independent proposals already surface many useful fragments, so spending more calls on lineage-conditioned rewrites gives little advantage over resampling. But once those fragments are retained, a set-level estimator can combine them globally and reject many interpolation-only supports. The same account explains why generic primitives are strong but incomplete: they provide a powerful mathematical prior, while LLM terms add problem-compatible scientific material.

\paragraph{Why per-term credit had to fail.}
The two-family separation is not a quirk of our credit formula; it is the LLM-era instance of a classical fact. LLM-generated dictionaries are highly correlated---a hundred proposals for one problem share structure by construction---and the training grid is small, so the design matrix is underdetermined. Under correlated, underdetermined designs, marginal screening of individual predictors is exactly the regime where classical sparse-recovery theory warns of unreliable support identification~\citep{fanlv2008,zhaoyu2006}, while the joint fit of a candidate support remains what the data determines. D3 measures this directly (per-term credit tracks functional contribution, Spearman $0.26$--$0.63$, yet truth-identification stays at chance), and the selector experiment converts it into performance: both per-term reductions forfeit roughly half the solves.

\paragraph{Why the program-domain result matters.}
A natural objection is that equations are unusually friendly to one-shot recombination because additive terms can be merged by least squares. The program-domain audit is included to test that objection without pretending that \method{} is a free-form program synthesizer. Online bin-packing programs can in principle be patched across generations; if the parent-lineage story were broadly valid, this domain should be favorable to it. Easy instances still reproduce the equation-domain null, and the harder DeepSeek signal attaches to retained external state and a crossover probe rather than to parent-conditioned generation count. The result is therefore a stress test of the evolutionary interpretation, not a negative result for the componentized method.

\paragraph{Limitations.}
(1) The strongest conclusion is about componentized scientific expression discovery and the common parent-conditioned propose--select loop, not every possible future LLM discovery system. (2) The official equation benchmark uses two proposers---Llama-3.1-8B and a single-seed DeepSeek-V4 anchor. A still stronger frontier proposer could change dictionary availability; the current cross-backbone gap, however, is small enough to make selection a load-bearing factor rather than a detail. (3) Symbolic accuracy is reported only as a caveated diagnostic, because a sparse additive readout can be numerically correct while not matching a compact closed form. We therefore make numeric and OOD claims, not an unconditional symbolic-SOTA claim. (4) LSR-Transform remains the weakest family, and the failure-mode autopsy attributes this mainly to dictionary coverage; better scientific proposal generators may help more than new selectors there. (5) The program-domain evidence is intentionally scoped: it is a small harder grid, not a full proof about all program synthesis or all proposers. Its role is to stress-test the evolutionary interpretation, not to extend \method{} as a general free-form program-synthesis method.

\paragraph{Future work.}
The immediate extension is not to add arbitrary benchmarks, but to move across componentized discovery spaces where the same design should apply: dynamical-system term discovery, PDE term selection, physics-inspired formula libraries, symbolic-regression suites beyond LLM-SRBench, and formula-like heuristic scoring rules. These tasks differ in domain but share the load-bearing structure: an LLM can propose reusable material, while data identifies sets of components better than lineage histories or individual-term credit. The hypothesis to carry forward is simple and falsifiable: when the bottleneck is material plus selection, one-shot proposal banks and set-level recombination should dominate parent-conditioned evolution; when a domain lacks a flat component bank, \method{} should be treated as out of scope rather than stretched into a universal program-synthesis recipe.

\section{Conclusion}

LLM scientific discovery is often described as evolution: keep the winners, mutate them, and let generations compound. This paper tests that description rather than assuming it. In continuous equation discovery, the parent-conditioned loop does not beat matched-budget sampling; it mainly produces a dictionary of candidate scientific terms. Built around the mechanism that survives the audit, \method{} reaches $73.2\%$ and a single-seed $77.0\%$ \acc{} on the official benchmark across a weak and a strong proposer at one tenth of the standardized budget. The program stress test shows that moving outside flat expression dictionaries does not rescue parent-conditioned generation count as the operative explanation. The broader lesson is that LLM discovery systems should be described by their actual division of labor: proposal material, retained state, verifier feedback, and set-level selection. Wherever an LLM proposes decomposable material and data must decide, decide at the set level.

\paragraph{Reproducibility.} All experiments carry pre-registered decision branches, frozen configs with hashes, blind-split evaluation, budget-fairness declarations, and leakage audits. Code, dictionaries, per-cell results, and audit artifacts will be released at \url{https://github.com/panli0/ptb-search}.

\bibliographystyle{iclr2026_conference}
\bibliography{refs}

\appendix

\section{Iteration schemes and lineage tracking}
E4 (naive re-extraction), M3 (credit-filtered heritable recombination), and R2 (guarded population-level recombination) are audit appendices characterizing the iteration regularity of \S\ref{sec:discussion}. Across the three schemes, the maximum strict-solve count is reached at generation 0 or remains tied with it; naive iteration on the development split degrades from 50 to 38 strict solves; the credit-filtered scheme inflates median formula length from 270 to 523 without OOD gain. Heritability and early-stopping ledgers are included in the release directories and summarized by Fig.~\ref{fig:generations}.

\section{Additional ablations and local references}
\label{app:additional-ablations}
Table~\ref{tab:additional-ablations} records the rows most relevant for interpreting \method{} against non-LLM or weakly-LLM references. PySR rows are local CPU references under our protocol, not official LLM-SRBench rows.

\begin{table}[h]
\centering\small
\caption{Additional local references. PySR and dense-dictionary rows are protocol context for failure-mode interpretation, not cross-proposer paired tests.}
\label{tab:additional-ablations}
\begin{tabular}{llccc}
\toprule
Group & Arm & Median OOD & \near{} & \solve{} \\
\midrule
DeepSeek dev & LLM terms only & $3.50\times10^{-6}$ & 46 & 36 \\
DeepSeek dev & Generic primitives & $2.69\times10^{-5}$ & 49 & 35 \\
DeepSeek dev & Dense (no subset selection) & $1.07\times10^{-7}$ & 55 & 40 \\
DeepSeek dev & Single proposal only & $2.19\times10^{-2}$ & 23 & 17 \\
DeepSeek dev & Local edits only & $6.00\times10^{-2}$ & 26 & 15 \\
DeepSeek dev & PySR matched & $1.07$ & 3 & 1 \\
DeepSeek dev & PySR generous & $1.91\times10^{-1}$ & 20 & 16 \\
DeepSeek blind & LLM terms only & $4.86\times10^{-3}$ & 35 & 15 \\
DeepSeek blind & Generic primitives & $1.92\times10^{-3}$ & 39 & 24 \\
DeepSeek blind & Dense (no subset selection) & $2.14\times10^{-3}$ & 37 & 19 \\
DeepSeek blind & Single proposal only & $1.03\times10^{-1}$ & 18 & 5 \\
DeepSeek blind & Local edits only & $1.19\times10^{-2}$ & 29 & 13 \\
DeepSeek blind & PySR matched & $1.74\times10^{-1}$ & 5 & 2 \\
Llama full & PySR matched & $3.16\times10^{-1}$ & 36 & 12 \\
\bottomrule
\end{tabular}
\end{table}

\section{Official-baseline composites and the DeepSeek anchor}
\label{app:official-composites}
The Llama-3.1-8B baseline rows in Table~\ref{tab:official} are all-239 composites weighted by problem count over the official Table-1 per-domain entries (LSR-Transform $111$; chemistry $36$, biology $24$, physics $43$, material science $25$). Each composite reproduces the official numbers exactly: LLM-SR \acc{} $=(38.55{\cdot}111 + 66.66{\cdot}36 + 58.33{\cdot}24 + 34.09{\cdot}43 + 88.12{\cdot}25)/239 = 49.15\%$; SA $=19.70\%$; LaSR \acc{} $=38.41\%$, SA $=4.61\%$; SGA \acc{} $=3.35\%$, SA $=1.25\%$; Direct \acc{} $=0.84\%$. We do not composite the baseline NMSE column, because the official table reports NMSE only per domain and a problem-count mean of per-domain NMSEs is not the all-239 \emph{median} our aggregation uses; the median-NMSE column of Table~\ref{tab:official} therefore carries only our own runs, each computed under the official Table-1 median aggregation.

Table~\ref{tab:deepseek-official} gives the DeepSeek-V4 anchor by family. The all-239 ID \acc{} is $77.0\%$; on the 128 LSR-Synth problems it reaches $96.1\%$ ID and $73.4\%$ OOD \acc{}, with material science again essentially solved. LSR-Transform has no OOD field in the loaded official datamodule, so its OOD entry is absent by construction rather than missing.

\begin{table}[h]
\centering\small
\caption{DeepSeek-V4 official anchor (seed 0, $23{,}900$ fresh proposals, frozen \method{}). ID and OOD are official \acc{}; medians are official NMSE. LSR-Transform supplies no OOD split in the official datamodule, so the all-239 OOD cell (\ddag) averages in Transform problems that have no OOD samples and understates OOD performance; the clean OOD figure is the LSR-Synth row.}
\label{tab:deepseek-official}
\begin{tabular}{lccccc}
\toprule
Scope & $n$ & ID \acc{} (\%) & OOD \acc{} (\%) & Median ID NMSE & Median OOD NMSE \\
\midrule
All 239            & 239 & 77.0 & 39.3\,\textsuperscript{\ddag} & $9.24\times10^{-5}$ & $2.09\times10^{-5}$ \\
LSR-Synth          & 128 & 96.1 & 73.4 & $6.58\times10^{-7}$ & $2.09\times10^{-5}$ \\
LSR-Transform      & 111 & 55.0 & --- & $5.00\times10^{-2}$ & --- \\
\midrule
Biology            &  24 & 91.7 & 70.8 & $1.46\times10^{-8}$ & $4.90\times10^{-4}$ \\
Chemistry          &  36 & 100.0 & 69.4 & $9.34\times10^{-8}$ & $2.12\times10^{-4}$ \\
Material science   &  25 & 96.0 & 88.0 & $8.18\times10^{-9}$ & $4.26\times10^{-7}$ \\
Physics            &  43 & 95.3 & 69.8 & $1.60\times10^{-5}$ & $3.77\times10^{-5}$ \\
\bottomrule
\end{tabular}
\end{table}

\paragraph{DeepSeek internal arms and audit checks.}
Table~\ref{tab:deepseek-core-arms} reports the frozen seed-0 official anchor before official aggregation. The same structure seen in development reappears on all 239 official problems: fresh proposals are much weaker; LLM-only and generic-only dictionaries both help; and the hybrid dictionary gives the best median OOD and the largest near/strict-solve counts in this anchor. Table~\ref{tab:deepseek-audit-checks} records the completeness and leakage checks for this anchor.

\begin{table}[h]
\centering\scriptsize
\caption{DeepSeek-V4 seed-0 official anchor: internal PTB arms on 239 cells. OOD is post-hoc; construction and selection are train-only.}
\label{tab:deepseek-core-arms}
\resizebox{\linewidth}{!}{
\begin{tabular}{lcccccccc}
\toprule
Arm & Median train & Median OOD & \near{} & \solve{} & Validity & Multi-root & Overfit rate & Median OOD lift \\
\midrule
Fresh100 & $1.98\times10^{-3}$ & $8.63\times10^{-2}$ & 49 & 21 & 0.999 & 0.000 & 0.000 & --- \\
LLM terms only & $2.56\times10^{-4}$ & $1.89\times10^{-2}$ & 81 & 36 & 1.000 & 0.950 & 0.100 & 0.238 \\
GenericPrimitiveSparse & $5.51\times10^{-4}$ & $5.25\times10^{-2}$ & 82 & 53 & 1.000 & 0.950 & 0.067 & 0.385 \\
Hybrid \method{} & $\mathbf{5.22\times10^{-5}}$ & $\mathbf{4.57\times10^{-3}}$ & \textbf{103} & \textbf{63} & 1.000 & 0.971 & 0.084 & 0.551 \\
\bottomrule
\end{tabular}}
\end{table}

\begin{table}[h]
\centering\small
\caption{DeepSeek-V4 seed-0 official anchor: completeness and leakage checks.}
\label{tab:deepseek-audit-checks}
\begin{tabular}{lc@{\qquad}lc}
\toprule
Check & Value & Check & Value \\
\midrule
Expected problems & 239 & Observed problems & 239 \\
Expected seed count & 1 & Observed seed count & 1 \\
Expected cells & 239 & Observed cells & 239 \\
Expected fresh attempts & 23{,}900 & Observed fresh attempts & 23{,}900 \\
Cells not exactly Fresh100 & 0 & Fresh valid rate & 0.999038 \\
Candidate-ID duplicates & 0 & Scoring errors & 0 \\
Uses OOD for fit/selection & False & Uses true formula/oracle for ranking & False \\
\bottomrule
\end{tabular}
\end{table}

\paragraph{Symbolic-readout diagnostic.}
The native sparse readout is deliberately not a compact closed form, so SA is not used as a headline. A train-only simplification pass modestly improves replacement-judge SA while preserving the numeric pattern; because the official GPT-4o judge was unavailable for this diagnostic run, we report this only as a caveat.

\begin{table}[h]
\centering\small
\caption{Train-only symbolic simplification diagnostic on the official split. The judge is a DeepSeek-V4-Pro replacement judge; no test/OOD/true formula is read during simplification.}
\label{tab:sa-simplification}
\begin{tabular}{lccccc}
\toprule
Readout & SA (\%) & Median terms & Mean removed & Median train NMSE & ID \acc{} \\
\midrule
Original frozen readout & 1.12 & 6 & --- & $4.74\times10^{-4}$ & 73.2 \\
Train-only simplified readout & 2.23 & 5 & 1.09 & $5.78\times10^{-4}$ & 73.6 \\
\bottomrule
\end{tabular}
\end{table}

\section{Generic baselines, complexity, and validity}
\label{app:generic-complexity}

Table~\ref{tab:llama-generic} reports the most important defensive comparison on the Llama full benchmark. Hybrid \method{} has the best median OOD NMSE, but GenericPrimitiveSparse has more \near{} and \solve{} cells. This is why we frame LLM-derived terms as complementary, problem-compatible priors rather than as the sole source of the gain.

\begin{table}[h]
\centering\small
\caption{Llama 717-cell full benchmark: generic sparse search is a strong baseline. Hybrid \method{} gives the best median OOD, while GenericPrimitiveSparse gives more \near{} and strict-solve cells.}
\label{tab:llama-generic}
\begin{tabular}{lcccccc}
\toprule
Arm & Median train & Median OOD & \near{} & \solve{} & Validity & Median length \\
\midrule
Fresh100 & $3.48\times10^{-2}$ & $4.23\times10^{-1}$ & 28 & 5 & 0.5835 & --- \\
LLM terms only & $4.17\times10^{-3}$ & $7.86\times10^{-2}$ & 179 & 76 & 1.0 & 268 \\
GenericPrimitiveSparse & $5.51\times10^{-4}$ & $5.15\times10^{-2}$ & \textbf{244} & \textbf{160} & 1.0 & \textbf{197} \\
Hybrid \method{} & $7.16\times10^{-4}$ & \textbf{$2.75\times10^{-2}$} & 239 & 135 & 1.0 & 236 \\
Single-root control & $2.94\times10^{-2}$ & $2.49\times10^{-1}$ & 79 & 28 & 1.0 & 119 \\
Cross-family control & $4.01\times10^{-2}$ & $2.99\times10^{-1}$ & 9 & 0 & 0.2469 & 272 \\
\bottomrule
\end{tabular}
\end{table}

\begin{figure}[!ht]
  \centering
  \makebox[\linewidth][c]{\includegraphics[width=1.23\linewidth]{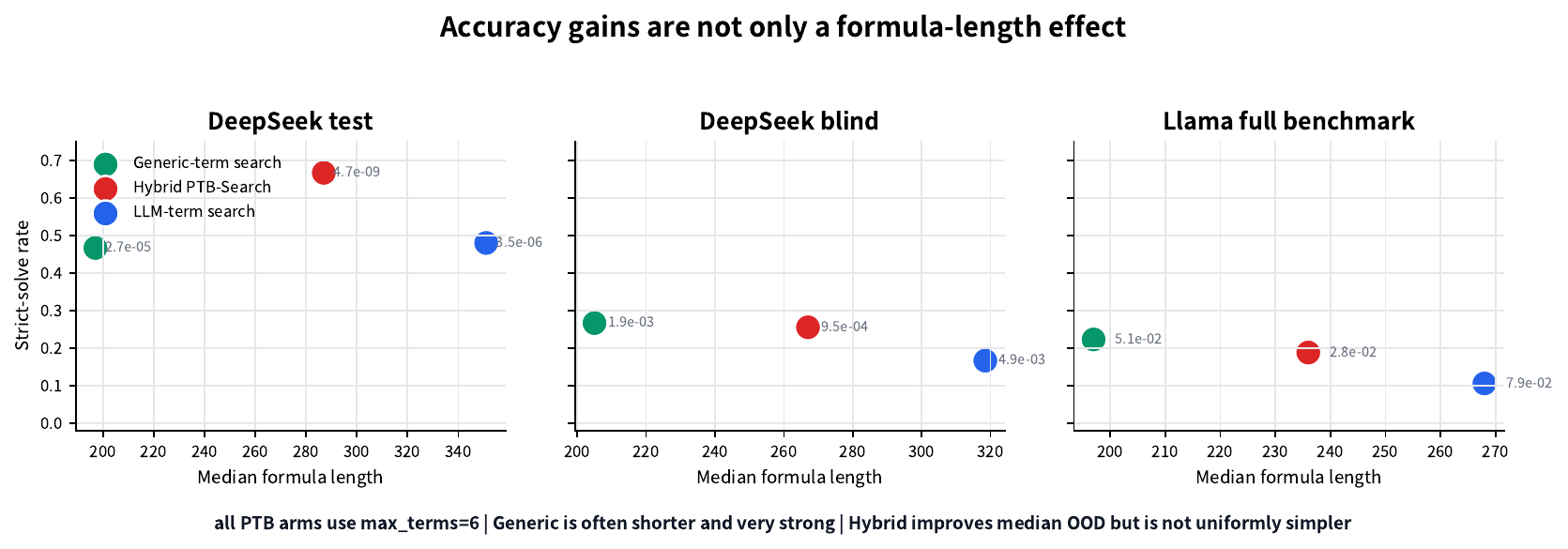}}
  \caption{\textbf{Accuracy is not bought by unconstrained formula length.} All PTB arms use max\_terms=6; generic terms are often shorter and strong, while hybrid dictionaries improve median OOD but are not uniformly simpler.}
  \label{fig:complexity-app}
\end{figure}

All PTB arms use the same sparse-term cap (six terms), so the gains are not an unconstrained length effect. Across DeepSeek test, DeepSeek blind, and Llama full benchmark, GenericPrimitiveSparse is often shorter and very strong; Hybrid is usually shorter than LLM-only and gives the best median OOD, but it is not uniformly simpler than generic-only. Figure~\ref{fig:complexity-app} visualizes this tradeoff.

\begin{table}[h]
\centering\scriptsize
\caption{Formula-complexity readout. All rows use the same sparse cap of six selected terms; the table reports median formula length, median selected terms, median OOD NMSE, and threshold counts.}
\label{tab:formula-complexity}
\resizebox{\linewidth}{!}{
\begin{tabular}{llccccc}
\toprule
Group & Arm & Median length & Median terms & Median OOD & \near{} & \solve{} \\
\midrule
DeepSeek dev & LLM terms only & 351 & 6 & $3.50\times10^{-6}$ & 46 & 36 \\
DeepSeek dev & GenericPrimitiveSparse & 197 & 6 & $2.69\times10^{-5}$ & 49 & 35 \\
DeepSeek dev & Hybrid \method{} & 287 & 6 & $\mathbf{4.70\times10^{-9}}$ & \textbf{57} & \textbf{50} \\
DeepSeek blind & LLM terms only & 318.5 & 6 & $4.86\times10^{-3}$ & 35 & 15 \\
DeepSeek blind & GenericPrimitiveSparse & 205 & 6 & $1.92\times10^{-3}$ & 39 & \textbf{24} \\
DeepSeek blind & Hybrid \method{} & 267 & 6 & $\mathbf{9.50\times10^{-4}}$ & \textbf{45} & 23 \\
Llama full & LLM terms only & 268 & 6 & $7.86\times10^{-2}$ & 179 & 76 \\
Llama full & GenericPrimitiveSparse & 197 & 6 & $5.15\times10^{-2}$ & \textbf{244} & \textbf{160} \\
Llama full & Hybrid \method{} & 236 & 6 & $\mathbf{2.75\times10^{-2}}$ & 239 & 135 \\
\bottomrule
\end{tabular}}
\end{table}

Local Llama proposals have lower raw parse/fit validity than DeepSeek (fresh100 validity $0.5835$), but PTB candidates are controlled sparse formulas built from valid extracted terms. The Llama fullbench generated exactly $71{,}700$ fresh attempts over 239 problems $\times$ three seeds, with exactly 100 attempts per cell, no duplicate candidate IDs, and no OOD/test/true/oracle access during fit, selection, ranking, or dictionary construction. PTB validity therefore reflects controlled symbolic recombination, not hidden LLM retries or OOD-aware filtering.

\section{Source-count ablation and BPG10 support table}
\label{app:source-bpg10}

A broader proposal-source pool matters: on the DeepSeek held-out split, restricting LLM-term extraction to the top 5 roots gives median OOD $7.98\times10^{-3}$, 26 \near{} cells, and 19 strict solves; top 20 gives $3.50\times10^{-4}$, 38, and 25; top 40 gives $8.02\times10^{-6}$, 44, and 32; the main bank gives $3.50\times10^{-6}$, 46, and 36. The gain is therefore not a lucky single-root effect.

The interpretability figure uses BPG10, seed 0. The selected candidate uses 80 fresh roots to build a dictionary with 49 LLM terms, 27 generic terms, and 74 hybrid terms; the selected formula has recomputed train NMSE 0 and OOD NMSE $2\times10^{-10}$. The winning support combines three LLM-root terms with two generic terms:
\[
\beta_0 + \beta_1 t^{1.131} + \beta_2 e^{-0.192t} - \beta_3(0.143P) + \beta_4 t^2P + \beta_5\sqrt{|P|+10^{-8}}.
\]
The selected terms come from multiple source roots plus generic primitives, so no single proposal held the final support.

\begin{table}[h]
\centering\small
\caption{BPG10 selected support terms used in Fig.~\ref{fig:interpretability}. OOD is reported post-hoc only.}
\label{tab:bpg10-support}
\begin{tabular}{ccccc}
\toprule
Term & Source & Source rank & Coefficient & Expression \\
\midrule
4  & LLM root & 1 & $0.00229$ & $t^{1.131192}$ \\
10 & LLM root & 1 & $-0.00842$ & $\exp(-0.192313t)$ \\
26 & LLM root & 6 & $-0.7044$ & $-0.142976P$ \\
59 & Generic & --- & $3.293\times10^{-8}$ & $t^2P$ \\
68 & Generic & --- & $0.02807$ & $\sqrt{|P|+10^{-8}}$ \\
\bottomrule
\end{tabular}
\end{table}

Table~\ref{tab:bpg10-top-supports} shows the candidate-support valley underlying Fig.~\ref{fig:interpretability}: several supports are nearly tied by the train/internal score, while their post-hoc OOD behavior can separate. This is why the paper treats individual-term credit as insufficient and scores supports jointly.

\begin{table}[h]
\centering\scriptsize
\caption{BPG10 top candidate supports. Stable score and split-select NMSE are train-side quantities; OOD NMSE is post-hoc only.}
\label{tab:bpg10-top-supports}
\resizebox{\linewidth}{!}{
\begin{tabular}{ccccccccc}
\toprule
Rank & Selected & Method & Term IDs & Terms & Stable score & Split-select NMSE & OOD NMSE & Source roots \\
\midrule
1 & 1 & BeamSparse8 & [4,10,26,59,68] & 5 & -28.50 & $4.21\times10^{-13}$ & $2\times10^{-10}$ & 4 \\
2 & 0 & BeamSparse8 & [8,10,26,59,68] & 5 & -28.50 & $4.21\times10^{-13}$ & $2\times10^{-10}$ & 4 \\
3 & 0 & BeamSparse8 & [4,10,18,26,68] & 5 & -28.09 & $6.33\times10^{-13}$ & $0$ & 4 \\
4 & 0 & BeamSparse8 & [8,10,26,56,68] & 5 & -28.09 & $6.33\times10^{-13}$ & $0$ & 4 \\
5 & 0 & BeamSparse8 & [4,10,26,56,68] & 5 & -28.09 & $6.33\times10^{-13}$ & $0$ & 4 \\
\bottomrule
\end{tabular}}
\end{table}

\section{Program-domain details}
\label{app:program-details}

The hard program-domain grid is a quick clean readout over seeds 0--2 only; seeds 3--4 were excluded after billing/API contamination rather than repaired. The per-distribution means are shown in Table~\ref{tab:program-per-dist}. We therefore use the result as a scoped positive boundary signal, not as a full program-domain proof.

\begin{table}[h]
\centering\small
\caption{Hard program-domain clean grid over three distributions $\times$ three seeds. Blind excess-bin rate is lower better.}
\label{tab:program-main}
\begin{tabular}{lcccc}
\toprule
Arm & Validity & Mean blind excess & Median blind excess & Mean blind gain vs best-fit \\
\midrule
Fresh $N=400$ & 0.997 & 3.907 & 3.879 & 0.372 \\
Parent evolution, 20 gen & 0.992 & 3.568 & 3.482 & 0.696 \\
External population, 20 gen & 0.998 & 3.361 & 3.273 & 0.895 \\
Crossover probe, 100 calls & 0.993 & \textbf{3.195} & \textbf{3.184} & \textbf{1.054} \\
\bottomrule
\end{tabular}
\end{table}

\begin{table}[h]
\centering\small
\caption{Hard program-domain paired comparisons against fresh sampling. Ties are ignored in the two-sided sign test.}
\label{tab:program-paired}
\begin{tabular}{lcccc}
\toprule
Comparison & Fresh minus arm & Wins & Losses & Sign-test $p$ \\
\midrule
Parent evolution vs fresh & 0.339 & 5 & 3 & 0.727 \\
External population vs fresh & 0.546 & 7 & 0 & 0.016 \\
Crossover probe vs fresh & 0.712 & 8 & 0 & 0.008 \\
\bottomrule
\end{tabular}
\end{table}

\begin{table}[h]
\centering\small
\caption{Hard program-domain per-distribution readout. Blind excess-bin rate is lower better.}
\label{tab:program-per-dist}
\begin{tabular}{llcc}
\toprule
Distribution & Arm & Mean blind excess (\%) & Median blind excess (\%) \\
\midrule
OR-style & Fresh & 3.637 & 3.795 \\
OR-style & Parent evolution & 3.178 & 3.168 \\
OR-style & External population & 3.276 & 3.287 \\
OR-style & Crossover probe & \textbf{3.068} & 3.287 \\
Tight & Fresh & 3.800 & 3.879 \\
Tight & Parent evolution & 3.712 & 3.715 \\
Tight & External population & 3.445 & 3.273 \\
Tight & Crossover probe & \textbf{3.225} & \textbf{3.184} \\
Weibull & Fresh & 4.283 & 4.295 \\
Weibull & Parent evolution & 3.814 & 3.635 \\
Weibull & External population & 3.361 & \textbf{3.166} \\
Weibull & Crossover probe & \textbf{3.292} & \textbf{3.166} \\
\bottomrule
\end{tabular}
\end{table}

\section{Audit artifacts}
\label{app:audit-artifacts}
The release artifact contains one result directory per experiment with decision logs, leakage checks, frozen run configs, per-cell CSVs, and pre-registered branch texts; a pre-submission machine audit (16 checks passed) verifies selector-code leakage freedom, dictionary identity across selector arms, paired-statistics implementation, capacity-sweep prefix integrity, and the correspondence of every number in this paper to the consolidated CSVs. A reproducibility snapshot records git state, environment, model inventory, raw-completion inventories, and file hashes.

\end{document}